\documentclass[10pt,leqno]{amsart}
\usepackage{graphicx}
\usepackage{indentfirst}
\usepackage{amssymb,amsthm,amsmath}
\usepackage{booktabs,multirow,array,microtype}
\usepackage[table]{xcolor}
\usepackage{hyperref}
\usepackage{comment}

\definecolor{lightblue}{RGB}{238,245,255}
\hypersetup{colorlinks=true,linkcolor=black,citecolor=black,filecolor=black,urlcolor=black}
\excludecomment{CCSXML}
\newcommand{\ccsdesc}[2][]{}
\newcommand{\mytablefont}{\fontsize{8.3pt}{9.6pt}\selectfont}

\AtBeginDocument{%
  }

\begin{document}

\title[SyncBreaker]{SyncBreaker: Stage-Aware Multimodal Adversarial Attacks on Audio-Driven Talking Head Generation}

\author[Zhang et al.]{%
Wenli Zhang$^1$\hspace{1em}
Xianglong Shi$^1$\hspace{1em}
Sirui Zhao$^{1,*}$\hspace{1em}
Xinqi Chen$^1$, \\
Guo Cheng$^2$\hspace{1em}
Yifan Xu$^1$\hspace{1em}
Tong Xu$^{1,*}$\hspace{1em}
Yong Liao$^1$}
\address{$^1$University of Science and Technology of China\\
$^2$Beijing University of Technology}
\email{siruit@ustc.edu.cn, tongxu@ustc.edu.cn}
\thanks{Corresponding authors: Sirui Zhao and Tong Xu.}
\begin{abstract}
Diffusion-based audio-driven talking-head generation enables realistic portrait animation, but also introduces risks of misuse, such as fraud and misinformation. Existing protection methods are largely limited to a single modality, and neither image-only nor audio-only attacks can effectively suppress speech-driven facial dynamics. To address this gap, we propose SyncBreaker, a stage-aware multimodal protection framework that jointly perturbs portrait and audio inputs under modality-specific perceptual constraints. Our key contributions are twofold. First, for the image stream, we introduce nullifying supervision with Multi-Interval Sampling (MIS) across diffusion stages to steer the generation toward the static reference portrait by aggregating guidance from multiple denoising intervals. Second, for the audio stream, we propose Cross-Attention Fooling (CAF), which suppresses interval-specific audio-conditioned cross-attention responses. Both streams are optimized independently and combined at inference time to enable flexible deployment. We evaluate SyncBreaker in a white-box proactive protection setting. Extensive experiments demonstrate that SyncBreaker more effectively degrades lip synchronization and facial dynamics than strong single-modality baselines, while preserving input perceptual quality and remaining robust under purification. 
Code: https://github.com/kitty384/SyncBreaker.

\end{abstract}


%


\maketitle

\section{Introduction}

Audio-driven talking-head generation animates a static portrait with a driving audio clip to produce a realistic speaking video. This technology has found broad applications in digital human, film production, and virtual assistants, among others. Recent advances in generative modeling~\cite{hallo3, loopy, vasa-1, emo} have significantly improved identity preservation, facial dynamics, and lip–speech alignment, pushing synthesized results toward unprecedented realism. 
The growing realism of talking-head synthesis, however, introduces new risks of misuse. 
Fabricated videos can be generated from a portrait image and audio clip, threatening individual privacy and public trust, especially in scenarios like deepfake-based fraud and misinformation. To counter such threats,  
developing proactive protection mechanisms is essential. One promising direction is to introduce adversarial perturbations to model inputs, which can disrupt the generation process and hinder malicious talking-head synthesis.

Mainstream talking-head generation systems are now predominantly built on diffusion architectures, conditioning on both a reference portrait and a driving audio clip. While adversarial protection has been explored for diffusion-based generative models~\cite{advdm,mist,photoguard,sds}, existing methods are largely designed for image generation or editing tasks. When applied to talking-head synthesis, they primarily degrade visual quality but fail to effectively suppress facial motion generation.
Silencer~\cite{silence} represents a notable effort targeting the reference portrait, aiming to induce static-mouth outputs. However, the driving audio still provides strong motion cues, so lip movements and other facial dynamics are often preserved. More importantly, most prior work focuses only on the visual input, i.e., the reference portrait, while paying little attention to the audio modality, even though audio is the primary driver of facial dynamics. Attacking audio is not straightforward either. Existing audio attacks~\cite{asr3,asr4,ASRAttacks,whisper,mute} are mainly developed for automatic speech recognition (ASR) and do not effectively interfere with the motion synthesis process in talking-head generation.
Consequently, no existing solution effectively disrupts the audio-driven motion synthesis process that lies at the heart of this task.
\begin{figure}[t]
    \centering
    \includegraphics[
        width=\columnwidth,
        keepaspectratio
    ]{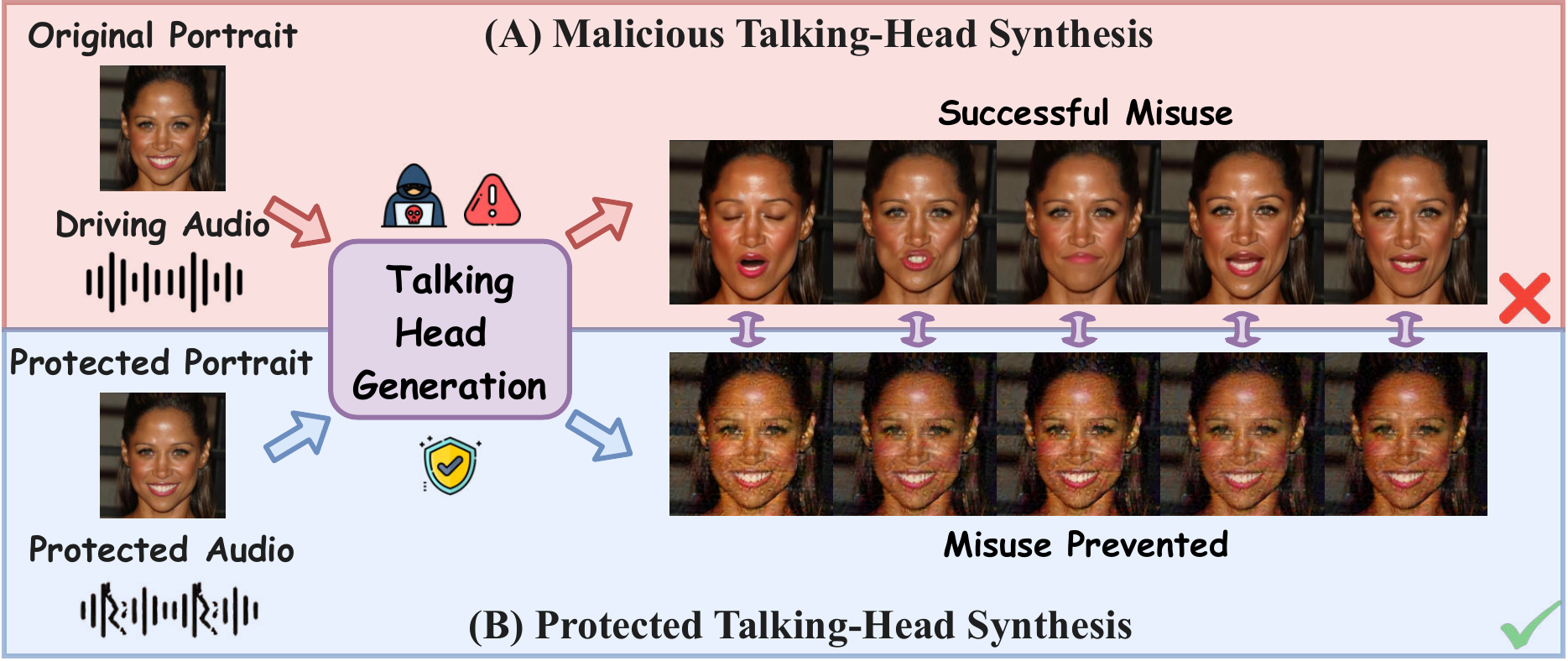}
    \caption{Overview of the proposed framework.}
    \label{fig:intro}
\end{figure}

To address these limitations, specifically the neglect of audio modality and the ineffectiveness of single-modal attacks, we propose SyncBreaker, a stage-aware multimodal adversarial attack framework for proactive protection against malicious talking-head synthesis. As illustrated in Fig.~\ref{fig:intro}, SyncBreaker applies separately optimized perturbations to the reference portrait and the driving audio, then feeds the protected inputs to the target generation model to disrupt facial motion synthesis.
Specifically, SyncBreaker decomposes multimodal protection into two coordinated streams. The image stream employs Multi-Interval Sampling (MIS)-based nullifying supervision, where timesteps are sampled from multiple diffusion-stage intervals to steer denoising toward a static reference portrait. The audio stream introduces Cross-Attention Fooling (CAF), which flattens audio-conditioned spatial responses by targeting interval-specific layer--branch unit sets, thereby weakening speech-to-motion guidance. 
The perturbations are optimized separately under modality-specific perceptual constraints and combined at inference time, destabilizing generated outputs and hindering the synthesis of facial dynamics while preserving input naturalness.

Our contributions are summarized as follows: 
\begin{itemize}
  \item We propose SyncBreaker, a novel stage-aware multimodal adversarial protection framework that reformulates proactive defense for audio-driven talking-head generation as coordinated perturbation learning over portrait and audio inputs. By jointly attacking both conditioning modalities, SyncBreaker effectively suppresses malicious synthesis while preserving input naturalness.

  \item  We develop two synergistic attack streams to disrupt generation. The image stream introduces a Multi-Interval Sampling (MIS)-based nullifying loss that aggregates supervision across denoising stages and steers outputs toward static reconstructions. In parallel, the audio stream employs Cross-Attention Fooling (CAF) to suppress interval-specific cross-attention responses.

  \item Extensive experiments on CelebA-HQ---LibriSpeech and HDTF demonstrate that SyncBreaker consistently outperforms strong image-only and audio-only baselines, substantially degrading lip synchronization and facial dynamics while maintaining high perceptual quality of protected inputs and strong robustness under purification defenses.
\end{itemize}

\section{Related Work}
\subsection{Audio-driven Talking-Head Generation}
Audio-driven talking-head generation has progressed rapidly, transitioning from intermediate motion representations to end-to-end generative models.  Early frameworks favored explicit motion modeling. ATVGNet~\cite{ATVGnet} was among the early works to adopt a cascaded framework from audio to keypoints and then to images, exploring the cross-modal mapping from speech to facial motion. MakeItTalk~\cite{MakeitTalk} achieves facial animation for arbitrary identities through landmark representations and identity disentanglement. ~\cite{pose} introduces external pose signals to enable pose-controllable talking-face generation. AD-NeRF~\cite{ad_nerf} introduces dynamic NeRF into this task to enhance the 3D representation capability. Subsequently, SadTalker~\cite{sadtalker} models facial expressions and head motions with 3D motion coefficients, while AniPortrait~\cite{aniportrait} combines 3D facial meshes, landmarks, and diffusion models to improve visual quality and temporal consistency.

With the development of diffusion models, end-to-end frameworks have gradually become an important research direction. DiffTalk~\cite{difftalk} and EMO~\cite{emo} are representative of this trend. They generate talking videos with diffusion models and reduce the reliance on explicit 3D modeling. Hallo~\cite{hallo} improves generation quality and stability through hierarchical audio-driven visual synthesis, and Hallo2~\cite{hallo2} further extends this line to long-duration and high-resolution scenarios. VASA-1~\cite{vasa-1} emphasizes high naturalness and real-time performance. Loopy~\cite{loopy} focuses on modeling long-term motion dependencies. LetsTalk~\cite{LetsTalk} employs a latent diffusion transformer to model audio-conditioned video generation, while FantasyTalking~\cite{fantasytalking} improves motion realism through a two-stage audio-visual alignment strategy and coherent motion synthesis. Sonic~\cite{sonic} emphasizes global audio perception and motion control, while ConsistTalk~\cite{ConsistTalk} focuses on temporal consistency in diffusion-based talking-head generation. In addition, EAT~\cite{EAT} and EdTalk~\cite{edtalk} improve the expressiveness and controllability of talking-head synthesis from the perspectives of emotion-controllable generation and disentangled modeling, respectively. In this work, we use Hallo as the pre-trained talking-head model.
\subsection{Adversarial Attacks}
\subsubsection{Adversarial Attacks in the Image Domain}
Adversarial attacks~\cite{pgd,fgsm,c&w,boost,trans-attack,patch,real-attack,fre-attack,trans2,color} in the image domain were originally developed to reveal the susceptibility of deep models to small input perturbations. More recently, similar ideas have been adopted for proactive protection against LDM-based editing and mimicry. Existing studies~\cite{photoguard, advdm, sds, mist} typically add imperceptible perturbations to input images to corrupt the conditioning cues extracted by diffusion models, thereby degrading downstream tasks such as image editing, style and content mimicry, and other image-conditioned generation tasks. These methods differ in both their optimization strategies and the components they target. AdvDM~\cite{advdm}, for example, generates adversarial examples by estimating gradients of the diffusion objective through Monte Carlo sampling over latent variables and maximizing the model loss to disrupt conditional generation. PhotoGuard~\cite{photoguard} protects images through encoder-level and diffusion-level attacks that manipulate latent representations and the denoising process. Mist~\cite{mist} combines semantic and textural losses to improve the transferability and robustness of protective perturbations across tasks. Diff-Protect~\cite{sds} incorporates score-distillation-based optimization into image protection and identifies the encoder as a key vulnerability in latent diffusion models.

Despite their effectiveness in image editing and image-conditioned generation, these methods are not specifically designed for audio-driven talking-head synthesis. In this setting, they tend to degrade visual quality without reliably disrupting speech-driven facial dynamics, especially lip motion. Silencer~\cite{silence} is one of the few methods proposed to address this problem. It introduces a two-stage portrait protection framework that combines a nullifying objective for suppressing audio-driven animation with a latent anti-purification mechanism for improved robustness. Nevertheless, suppression remains incomplete, and residual speech-correlated mouth dynamics are still observable in many cases.
\begin{figure*}[ht]
    \centering
    \includegraphics[width=\textwidth]{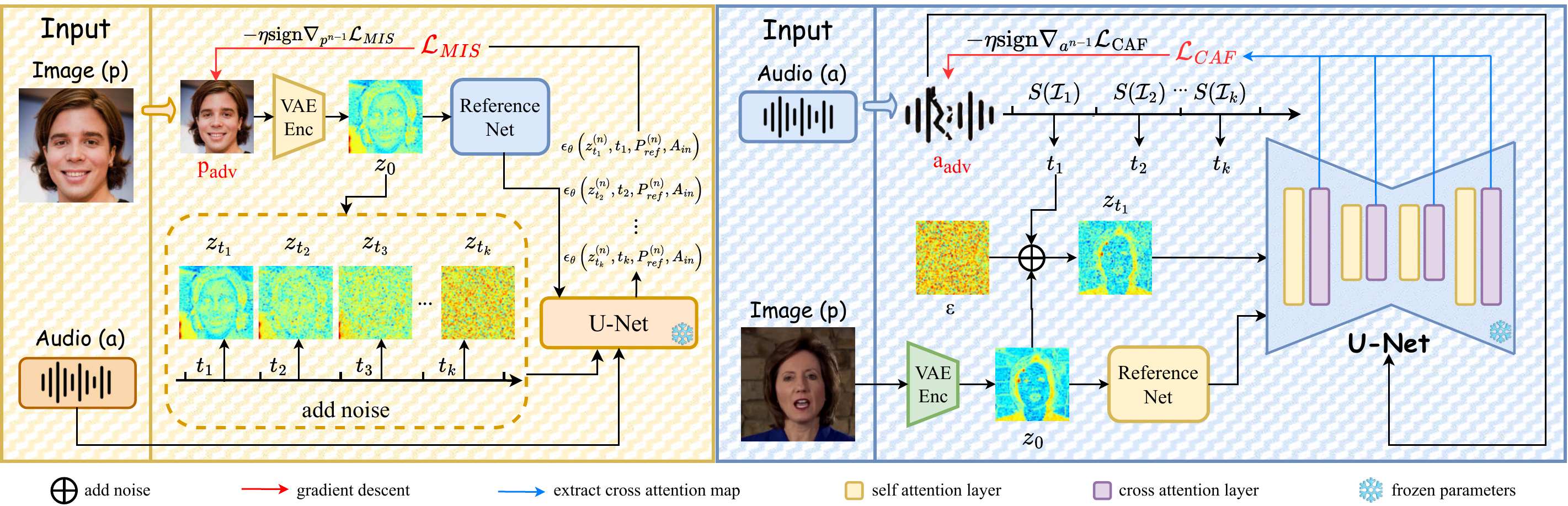}
    \caption{Overview of SyncBreaker. The image stream employs MIS-based Nullifying Loss to redirect the generation objective from synchronized speaking frames to a static reference image across multiple diffusion stages. The audio stream uses CAF to disrupt audio-visual cross-attention and break the alignment between the audio condition and key facial regions. The two streams are optimized separately and combined at inference time.}
    \label{fig:framework}
\end{figure*}

\subsubsection{Adversarial Attacks in the Audio Domain}
Existing audio adversarial attacks have mainly been studied in automatic speech recognition (ASR)~\cite{deepspeech,whisper_model}, where small perturbations are added to speech signals to cause recognition errors or attacker-specified transcriptions. Carlini and Wagner~\cite{asr4} were the first to systematically demonstrate targeted attacks on end-to-end speech recognition systems, showing that DeepSpeech~\cite{deepspeech} can be forced to output any desired phrase while keeping the adversarial audio highly similar to the original input. Qin et al.~\cite{qin} improved imperceptibility by incorporating psychoacoustic masking constraints and further enhanced robustness under physical playback by simulating environmental distortions. Du et al. proposed SirenAttack~\cite{asr3}, extending adversarial attacks to a broader class of end-to-end acoustic systems and demonstrating effectiveness as well as transferability in both white-box and black-box settings. As large-scale speech foundation models have emerged, recent work has also examined the adversarial vulnerability of newer ASR systems such as Whisper~\cite{whisper_model}. Olivier and Raj~\cite{whisper} found that although Whisper is relatively robust to random noise and distribution shifts, this robustness does not extend to adversarial perturbations: even small, carefully designed perturbations can substantially degrade recognition performance or induce target transcriptions. Raina et al. proposed Muting Whisper~\cite{mute}, which learns a universal short audio prefix that causes Whisper to emit the end-of-text token prematurely, thereby terminating transcription early across different inputs and tasks.
Despite their effectiveness, these methods are primarily designed for ASR and therefore do not adequately address the challenges of audio-driven talking-head generation, where the goal is not to alter linguistic transcription but to disrupt speech-driven facial motion.

\begin{figure*}[h]
    \centering
    \includegraphics[width=\textwidth]{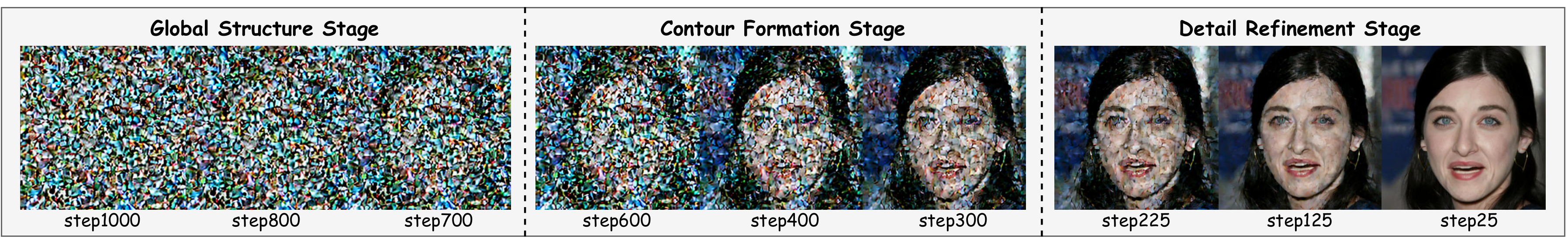}
    \caption{Visualization of denoising results across different diffusion stages, including the global structure stage, contour formation stage, and detail refinement stage.}
    \label{fig:step}
\end{figure*}
\begin{figure}[h]
    \centering
    \includegraphics[width=\linewidth]{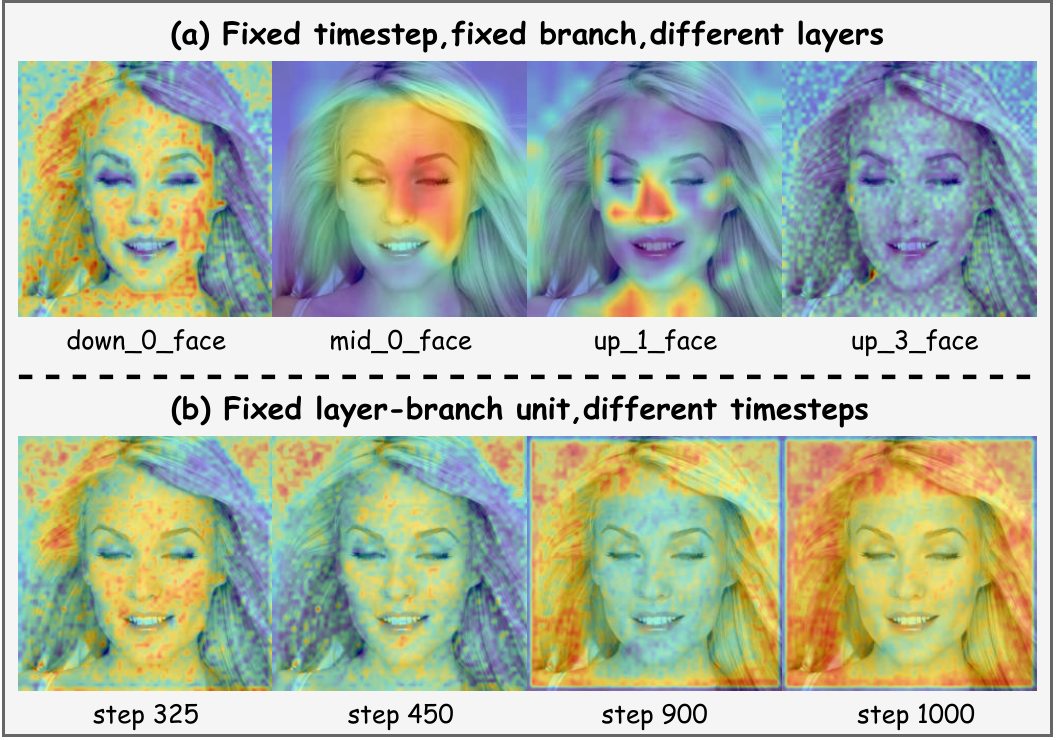}
    \caption{Audio-conditioned cross-attention maps. (a) At a fixed timestep and branch, different U-Net layers show distinct patterns. (b) For a fixed layer-branch unit, patterns vary across timesteps, with some similarity.}
    \label{fig:cross}
\end{figure}

\begin{figure*}[t]
    \centering
    \includegraphics[width=\textwidth]{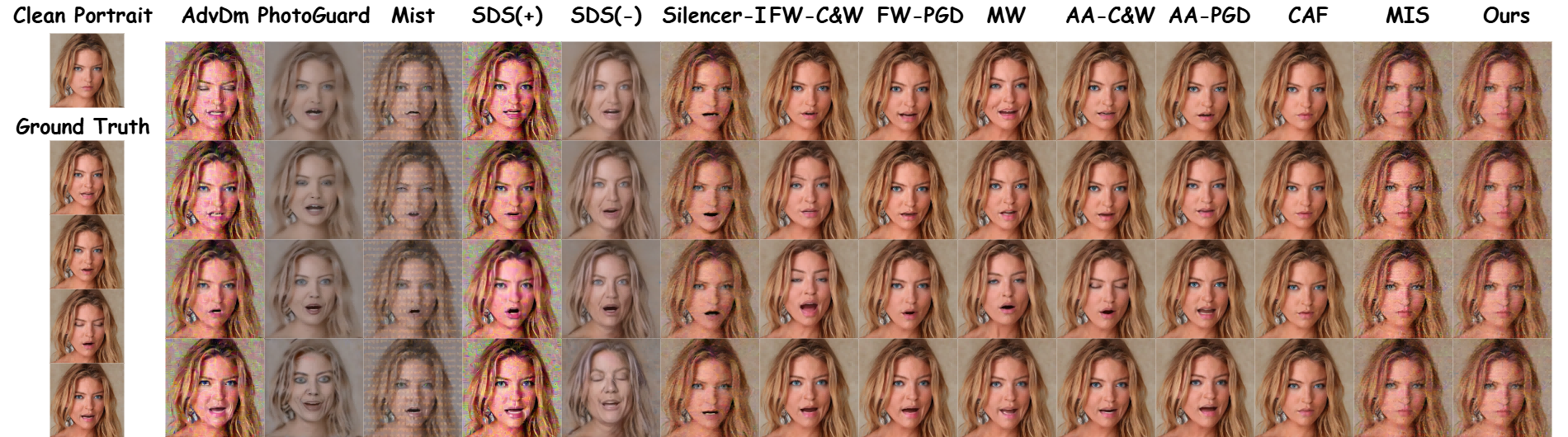}
    \caption{Qualitative comparison of videos generated from inputs protected by all compared attack methods.}
    \label{fig:all-qual}
\end{figure*}
\section{Method}
We present SyncBreaker, a multimodal proactive protection framework for diffusion-based talking-head generation. Fig.~\ref{fig:framework} illustrates how the proposed multimodal attack paradigm is instantiated in SyncBreaker. Specifically, the framework operates on both the reference image and the driving audio under modality-specific attack designs derived from the unified paradigm. In the following, we first define the multimodal attack paradigm, and then describe the two modality-specific methods.
\subsection{Multimodal Attack Paradigm}
\label{sec:attack_paradigm}

We consider a white-box proactive protection setting, where the defender has access to the architecture and parameters of the target talking-head generation model during perturbation optimization. Let $M$ denote the victim talking-head generation model, which takes a reference image $P_{ref}$ and driving audio $A_{in}$ as inputs and produces an output video $V_{out}$:
\begin{align}
    V_{out} = M(A_{in}, P_{ref}),
\end{align}

The goal of the multimodal attack is to introduce imperceptible perturbations into both the reference image and the driving audio so as to disrupt speech-driven facial dynamics in the generated video. Specifically, let $\delta_p$ and $\delta_a$ denote the perturbations added to the reference image and the driving audio, respectively. The perturbed inputs are defined as:
\begin{align}
    P'_{ref} &= P_{ref} + \delta_p,\\
    A'_{in}  &= A_{in} + \delta_a,
\end{align}
and the corresponding model output is:
\begin{align}
    V'_{out} = M(A'_{in}, P'_{ref}).
\end{align}

Under this formulation, the attack objective is to disrupt speech-driven facial dynamics while constraining perturbation magnitude in both modalities to preserve imperceptibility. Accordingly, the multimodal attack can be written as:
\begin{align}
    \min_{\delta_p,\delta_a}\quad
    \mathcal{L}_{adv}
    + \lambda_p \mathcal{R}_p(\delta_p)
    + \lambda_a \mathcal{R}_a(\delta_a),
\end{align}
where $\mathcal{L}_{adv}$ denotes the adversarial objective for disrupting speech-driven facial dynamics, $\mathcal{R}_p(\delta_p)$ and $\mathcal{R}_a(\delta_a)$ denote the constraints on image and audio perturbations, respectively, and $\lambda_p$ and $\lambda_a$ control the trade-off between attack effectiveness and imperceptibility.

In diffusion-based talking-head generation~\cite{hallo,hallo2,hallo3}, the reference image and the driving audio play fundamentally different roles: the former provides a static appearance prior for identity and visual consistency, whereas the latter supplies dynamic motion cues that drive speech-driven facial dynamics through cross-attention. These differences are difficult to capture with a single unified objective. Therefore, the multimodal attack is further instantiated as two modality-specific subproblems:
\begin{align}
    \delta_p^{*} &= \arg\min_{\delta_p}\; \mathcal{L}_{p}
    \quad \text{s.t. } \mathcal{R}_p(\delta_p) \leq \epsilon_p,\\
    \delta_a^{*} &= \arg\min_{\delta_a}\; \mathcal{L}_{a}
    \quad \text{s.t. } \mathcal{R}_a(\delta_a) \leq \epsilon_a.
\end{align}
Here, $\mathcal{L}_{p}$ and $\mathcal{L}_{a}$ denote the attack objectives for the image and audio modalities, respectively, and $\epsilon_p$ and $\epsilon_a$ are the corresponding perturbation budgets. Such a decomposition allows each modality-specific perturbation to maintain independent attack effectiveness, while also better matching practical dissemination scenarios in which portrait images and driving audio may be distributed or reused independently. In the full multimodal setting, the optimized perturbations $\delta_p^{}$ and $\delta_a^{}$ are jointly applied at inference time to disrupt speech-driven facial dynamics in the generated video.

\subsection{MIS-based Nullifying Loss}

In LDM-based talking-head generation, the reference image and driving audio jointly condition the denoising network to recover the result from noisy latent variables. Let $P_{ref}$ denote the reference image, $A_{in}$ the driving audio, and $\epsilon_\theta(\cdot)$ the denoising network.

In the proactive protection setting, the target speaking frame corresponding to the driving audio is unavailable. Consequently, image perturbation optimization cannot rely on ground-truth supervision as in standard diffusion training. Instead, nullifying loss~\cite{silence} uses the reference image itself as a static recovery target, encouraging the denoising process to reconstruct a still portrait rather than generate audio-driven speaking motions.

Specifically, at the $n$-th iteration, the current protected reference image $P_{ref}^{(n)}$ is first encoded into the latent space:
\begin{align}
z_0^{(n)} = E\!\left(P_{ref}^{(n)}\right),
\end{align}
where $E(\cdot)$ denotes the VAE encoder. Given a sampled timestep $t$, the forward diffusion process adds Gaussian noise $\epsilon \sim \mathcal{N}(0,I)$ to $z_0^{(n)}$, yielding:
\begin{align}
z_t^{(n)} = \sqrt{\bar{\alpha}_t}\, z_0^{(n)} + \sqrt{1-\bar{\alpha}_t}\,\epsilon,
\end{align}
where $\bar{\alpha}_t = \prod_{s=1}^{t}\alpha_s$ denotes the cumulative product of the diffusion noise schedule. The nullifying loss is then defined as:
\begin{align}
\mathcal{L}_{N}(t)
=
\left\|
\epsilon-\epsilon_\theta\!\left(z_t^{(n)},\, t,\, P_{ref}^{(n)},\, A_{in}\right)
\right\|_2^2,
\end{align}
Minimizing this loss drives the denoising trajectory away from audio-driven motions and toward the static reference portrait.

Furthermore, we observe that different denoising stages are responsible for recovering different types of visual content. As illustrated in Fig.~\ref{fig:step}, the early stages mainly determine the subject location, overall composition, and coarse structure, middle stages progressively establish clearer facial geometry and contours, and late stages further restore fine-grained textures and local visual details. These stage-wise differences suggest that different denoising stages capture complementary visual information.

However, Silencer~\cite{silence} samples only one timestep from a fixed interval during optimization, limiting supervision to a narrow stage of the denoising process.  To address this issue, we introduce a Multi-Interval Sampling (MIS) strategy, which samples timesteps from multiple intervals and applies nullifying supervision to leverage complementary information from different denoising stages.

Let $\{\mathcal{I}_k\}_{k=1}^{K}$ denote a set of timestep intervals. For each interval $\mathcal{I}_k$, we independently sample:
\begin{align}
t_k \sim \mathcal{U}(\mathcal{I}_k), 
\qquad
\epsilon_k \sim \mathcal{N}(0,I),
\qquad k=1,\dots,K,
\end{align}
and construct the corresponding noisy latent as:
\begin{align}
z_{t_k}^{(n)} = \sqrt{\bar{\alpha}_{t_k}}\, z_0^{(n)} + \sqrt{1-\bar{\alpha}_{t_k}}\,\epsilon_k,
\end{align}
The MIS objective for the image stream is given by:
\begin{align}
\mathcal{L}_{MIS}
=
\sum_{k=1}^{K}\lambda_k\,
\mathbb{E}_{\substack{t_k \sim \mathcal{U}(\mathcal{I}_k)\\ \epsilon_k \sim \mathcal{N}(0,I)}}
\left[
\left\|
\epsilon_k-\epsilon_\theta\!\left(z_{t_k}^{(n)},\, t_k,\, P_{ref}^{(n)},\, A_{in}\right)
\right\|_2^2
\right],
\end{align}
where $\lambda_k$ denotes the weight associated with the $k$-th timestep interval. During optimization, one timestep is sampled from each interval per iteration to compute nullifying supervision.

Compared with single-interval nullifying loss, MIS aggregates optimization signals from multiple denoising stages, enabling the perturbation to jointly influence global structure, facial contours, and fine details. This stronger stage-wise coverage improves the ability of the protected reference image to suppress audio-driven facial dynamics and steer generation toward a static portrait. 
Visually, this is typically reflected in weaker lip synchronization and reduced facial dynamics, including expression changes and blinking.

During optimization, we iteratively update the reference image using Projected Gradient Descent (PGD):
\begin{align}
P_{ref}^{(n+1)}
=
\Pi_{\mathcal{B}_\infty(P_{ref}^{(0)},\,\tau)}
\left(
P_{ref}^{(n)}
-\eta_p \cdot
\mathrm{sign}
\bigl(
\nabla_{P_{ref}^{(n)}} \mathcal{L}_{MIS}
\bigr)
\right),
\end{align}
where $\eta_p$ denotes the step size, $\tau$ denotes the perturbation budget, and $\Pi(\cdot)$ denotes the projection operator. Here, $\mathcal{B}_\infty(P_{ref}^{(0)},\,\tau)$ denotes the $L_\infty$ ball centered at the reference image $P_{ref}^{(0)}$ with radius $\tau$. 

\begin{table*}[t]
\centering
\small
\setlength{\tabcolsep}{2.5pt}
\renewcommand{\arraystretch}{0.74}
\caption{Quantitative comparisons with state-of-the-art methods on two test protocols: (1) CelebA-HQ images paired with LibriSpeech audio, and (2) HDTF dataset. Metrics marked with "$\uparrow$" indicate that higher values are better, while those marked with "$\downarrow$" indicate that lower values are better. Best results are highlighted in bold, while second-best results are underlined.}
\resizebox{0.92\textwidth}{!}{%
\begin{tabular}{l|c|ccccc|ccccc}
\toprule
\multirow{2}{*}{Method} & \multirow{2}{*}{Modality}
& \multicolumn{5}{c|}{CelebA-HQ --- LibriSpeech}
& \multicolumn{5}{c}{HDTF} \\
\cline{3-12}
&
& \rule{0pt}{2.2ex}\raisebox{-0.3ex}{V-PSNR$\downarrow$}
& \rule{0pt}{2.2ex}\raisebox{-0.3ex}{V-SSIM$\downarrow$}
& \rule{0pt}{2.2ex}\raisebox{-0.3ex}{FID$\uparrow$}
& \rule{0pt}{2.2ex}\raisebox{-0.3ex}{Sync$\downarrow$}
& \rule{0pt}{2.2ex}\raisebox{-0.3ex}{M-LMD$\uparrow$}
& \rule{0pt}{2.2ex}\raisebox{-0.3ex}{V-PSNR$\downarrow$}
& \rule{0pt}{2.2ex}\raisebox{-0.3ex}{V-SSIM$\downarrow$}
& \rule{0pt}{2.2ex}\raisebox{-0.3ex}{FID$\uparrow$}
& \rule{0pt}{2.2ex}\raisebox{-0.3ex}{Sync$\downarrow$}
& \rule{0pt}{2.2ex}\raisebox{-0.3ex}{M-LMD$\uparrow$} \\
\midrule
AdvDm~\cite{advdm}      & V  & 20.46 & \underline{0.42} & 181.90 & 5.33 & 4.03 & 21.39 & \underline{0.44} & \underline{215.68} & 5.81 & 3.12 \\
PhotoGuard~\cite{photoguard} & V  & \textbf{12.29} & 0.48 & 74.87  & 5.98 & 5.53 & \textbf{17.16} & 0.64 & 107.19 & 6.61 & 3.17 \\
Mist~\cite{mist}       & V  & 19.39 & 0.56 & \underline{209.44} & 4.87 & 4.50 & 21.04 & 0.61 & \textbf{256.21} & 4.78 & 3.36 \\
SDS(+)~\cite{sds}       & V  & 20.31 & \textbf{0.41} & 161.74 & 5.52 & 3.98 & 21.21 & \textbf{0.42} & 186.35 & 6.06 & 3.29 \\
SDS(-)~\cite{sds}       & V  & \underline{18.61} & 0.59 & 54.51  & 5.95 & 4.08 & \underline{20.04} & 0.66 & 79.80 & 6.35 & 2.99 \\
Silencer-I~\cite{silence} & V  & 21.86 & 0.50 & 176.32 & 3.30 & 5.46 & 24.69 & 0.62 & 166.92 & 3.16 & 3.43 \\
FW-C\&W ~\cite{whisper}   & A  & 23.15 & 0.74 & 5.78   & 3.63 & 4.26 & 35.66 & 0.94 & 1.86  & 6.64 & 1.17 \\
FW-PGD~\cite{whisper}     & A  & 25.09 & 0.78 & 4.62   & 5.20 & 3.23 & 31.21 & 0.91 & 2.56  & 5.89 & 1.79 \\
MW~\cite{mute}         & A  & 21.99 & 0.71 & 6.78   & 6.05 & 2.99 & 28.05 & 0.88 & 3.32  & 7.07 & 2.44 \\
AA-C\&W~\cite{ASRAttacks}    & A  & 25.67 & 0.79 & 4.25   & 5.25 & 2.88 & 33.59 & 0.93 & 1.96  & 6.50 & 1.38 \\
AA-PGD~\cite{ASRAttacks}     & A  & 24.39 & 0.77 & 4.75   & 3.70 & 3.97 & 31.65 & 0.92 & 2.41  & 5.50 & 1.86 \\
\rowcolor{lightblue}
\textbf{CAF}  & A  & 22.76 & 0.72 & 8.60   & \underline{1.85} & 4.60 & 29.31 & 0.89 & 3.69  & \underline{2.50} & 2.38 \\
\rowcolor{lightblue}
\textbf{MIS}  & V  & 20.05 & 0.46 & 203.96 & 2.82 & \underline{5.65} & 23.03 & 0.57 & 203.74 & 2.84 & \textbf{3.83} \\
\rowcolor{lightblue}
\textbf{Ours} & AV
              & 19.98
              & 0.46
              & \textbf{210.43}
              & \textbf{0.85}
              & \textbf{6.26}
              & 22.98
              & 0.56
              & 204.28
              & \textbf{1.07}
              & \underline{3.68} \\
\midrule
\textcolor{gray}{Ground Truth} & \textcolor{gray}{--} & \textcolor{gray}{$\infty$} & \textcolor{gray}{1} & \textcolor{gray}{--} & \textcolor{gray}{6.01} & \textcolor{gray}{0} & \textcolor{gray}{$\infty$} & \textcolor{gray}{1} & \textcolor{gray}{--} & \textcolor{gray}{6.96} & \textcolor{gray}{0} \\
\bottomrule
\end{tabular}%
}
\label{tab:all}
\end{table*}

\subsection{Cross-Attention Fooling}

Rather than altering audio semantics, CAF targets the injection path of the audio condition in the denoising network by weakening audio-conditioned cross-attention, thereby reducing the control of the audio signal over facial motion generation.

Hallo~\cite{hallo} injects audio conditions through cross-attention modules at multiple U-Net layers, where each injection location contains three branches: \textit{lip}, \textit{expression}, and \textit{pose}. We treat each layer-branch unit as a basic object for analyzing audio-conditioned cross-attention. As shown in Fig.~\ref{fig:cross}, the cross-attention responses vary across both U-Net layers and diffusion timesteps. Even within the same branch, different U-Net layers exhibit distinct spatial patterns, while for a fixed layer-branch unit, the response pattern changes over timesteps and remains similar over certain timestep ranges.  This suggests that audio-conditioned cross-attention has both layer-wise variation and stage-wise structure during denoising. Motivated by this observation, we partition the denoising process into multiple timestep intervals according to response-pattern similarity. Let $\{\mathcal{I}_k\}_{k=1}^{K}$ denote the set of timestep intervals. For each interval $\mathcal{I}_k$, we define a corresponding target layer-branch set $S(\mathcal{I}_k)$, where each unit $(\ell,b)$ denotes a cross-attention unit selected for that interval.

At the $n$-th iteration, we first randomly select a timestep interval $\mathcal{I}_k$ and sample a timestep:
\begin{align}
t \sim \mathcal{U}(\mathcal{I}_k),
\end{align}
Since no real speaking frame strictly corresponding to the driving audio is available in the proactive protection setting, we cannot obtain the noisy latent corresponding to the real generated result at timestep $t$. As in the nullifying loss, we use the current reference image $P_{ref}^{(n)}$ to construct the noisy latent input. The difference lies in the objective: the nullifying loss uses this latent to impose static nullifying supervision, whereas CAF uses it to probe and suppress audio-conditioned cross-attention responses. Specifically:
\begin{align}
z_0^{(n)} = E\!\left(P_{ref}^{(n)}\right), \qquad
z_t^{(n)} = \sqrt{\bar{\alpha}_t}\, z_0^{(n)} + \sqrt{1-\bar{\alpha}_t}\,\epsilon,
\end{align}
where $E(\cdot)$ denotes the VAE encoder and $\epsilon \sim \mathcal{N}(0,I)$. Using this latent, we extract the cross-attention maps produced at timestep $t$ by the layer-branch units in the target set $S(\mathcal{I}_k)$, denoted as:
\begin{align}
\left\{A_t^{(\ell,b)}\right\}_{(\ell,b)\in S(\mathcal{I}_k)},
\end{align}

When audio conditioning strongly influences facial motion, the corresponding cross-attention maps usually tend to be spatially concentrated on motion-relevant regions. To weaken this guidance effect, we reduce their spatial variance and define the CAF loss as:
\begin{align}
\mathcal{L}_{CAF}
=
\frac{1}{|S(\mathcal{I}_k)|}
\sum_{(\ell,b)\in S(\mathcal{I}_k)}
\mathrm{Var}\!\left(A_t^{(\ell,b)}\right),
\end{align}
where $\mathrm{Var}(\cdot)$ denotes the variance computed over the spatial elements of the corresponding attention map. Minimizing this loss drives the attention responses from highly concentrated distributions toward flatter spatial distributions, thereby weakening the alignment between audio features and facial motion regions, as well as the guidance of the audio condition over facial motion.

We do not jointly optimize multiple timesteps in each iteration. Instead, the audio is updated by randomly selecting one interval and sampling one timestep from it. This is because the cross-attention responses at all layers need to be retained during the denoising process, from which the target layer-branch units for the current timestep interval are selected for loss computation. Introducing multiple timesteps simultaneously would require retaining the cross-attention responses, gradient information, and computation graphs for all of them at once, resulting in substantial memory overhead. Random interval sampling therefore offers a more practical trade-off between attack effectiveness and optimization efficiency.

Finally, we iteratively update the input audio using PGD:
\begin{align}
A_{in}^{(n+1)}
=
\Pi_{\mathcal{C}_a}
\left(
A_{in}^{(n)}
-\eta_a \cdot
\mathrm{sign}
\bigl(
\nabla_{A_{in}^{(n)}} \mathcal{L}_{CAF}
\bigr)
\right),
\end{align}
where $\eta_a$ is the step size, $\Pi(\cdot)$ denotes the projection operator, and $\mathcal{C}_a$ denotes the feasible set determined by the distortion constraint. 

\begin{table}
\centering
\mytablefont
\setlength{\tabcolsep}{8pt}
\renewcommand{\arraystretch}{0.95}
\caption{Quantitative comparisons of adversarial image quality on CelebA-HQ and HDTF.}
\begin{tabular}{l|c|c}
\toprule
\multirow{2}{*}{Method} & CelebA-HQ & HDTF \\
\cline{2-3}
& \raisebox{-0.7ex}{I-PSNR/I-SSIM$\uparrow$} & \raisebox{-0.7ex}{I-PSNR/I-SSIM$\uparrow$} \\
\midrule
AdvDm~\cite{advdm}      & 27.30/0.59 & 27.29/0.56 \\
PhotoGuard~\cite{photoguard} & 27.29/0.57 & 27.41/0.55 \\
Mist~\cite{mist}       & 26.79/0.57 & 26.86/0.55 \\
SDS(+)~\cite{sds}       & 27.55/0.62 & 27.58/\underline{0.59} \\
SDS(-)~\cite{sds}       & 28.53/0.62 & 28.48/\underline{0.59} \\
Silencer-I~\cite{silence} & \textbf{29.91}/\textbf{0.70} & \textbf{29.96}/\textbf{0.66} \\
\rowcolor{lightblue}
\textbf{MIS} & \underline{29.56}/\underline{0.69} & \underline{29.59}/\textbf{0.66} \\
\bottomrule
\end{tabular}
\label{tab:image-ae}
\end{table}

\begin{table}
\centering
\mytablefont
\setlength{\tabcolsep}{8pt}
\renewcommand{\arraystretch}{0.95}
\caption{Quantitative comparisons of adversarial audio quality on LibriSpeech and HDTF.}
\begin{tabular}{l|c|c}
\toprule
\multirow{2}{*}{Method} & LibriSpeech & HDTF \\
\cline{2-3}
& \raisebox{-0.7ex}{SNR/PESQ$\uparrow$} & \raisebox{-0.7ex}{SNR/PESQ$\uparrow$} \\
\midrule
FW-C\&W~\cite{whisper} & 3.94/1.02 & 4.62/1.08 \\
FW-PGD~\cite{whisper}  & 17.22/1.21 & 18.11/1.57 \\
MW~\cite{mute}      & --/-- & --/-- \\
AA-C\&W~\cite{ASRAttacks} & \underline{22.40}/\underline{1.58} & \underline{19.30}/\underline{2.10} \\
AA-PGD~\cite{ASRAttacks}   & 1.08/1.02 & 5.37/1.08 \\
\rowcolor{lightblue}
\textbf{CAF} & \textbf{24.86}/\textbf{1.63} & \textbf{26.53}/\textbf{2.45} \\
\bottomrule
\end{tabular}
\label{tab:audio-ae}
\end{table}

\begin{table*}[t]
\centering
\mytablefont
\setlength{\tabcolsep}{1.7pt}
\renewcommand{\arraystretch}{1.1}
\caption{Quantitative comparison of different image attack methods under four purifiers. }
\resizebox{\textwidth}{!}{%
\begin{tabular}{l|cccc|cccc|cccc|cccc}
\toprule
\multirow{2}{*}{Method}
& \multicolumn{4}{c|}{JPEG~\cite{jpeg}}
& \multicolumn{4}{c|}{Resize~\cite{resize}}
& \multicolumn{4}{c|}{DiffPure~\cite{diffpure}}
& \multicolumn{4}{c}{DiffShortcut~\cite{DiffShortcut}} \\
\cline{2-17}
& \raisebox{-0.6ex}{I-PSNR/I-SSIM$\downarrow$} & \raisebox{-0.6ex}{FID$\uparrow$} & \raisebox{-0.6ex}{Sync$\downarrow$} & \raisebox{-0.6ex}{M-LMD$\uparrow$}
& \raisebox{-0.6ex}{I-PSNR/I-SSIM$\downarrow$} & \raisebox{-0.6ex}{FID$\uparrow$} & \raisebox{-0.6ex}{Sync$\downarrow$} & \raisebox{-0.6ex}{M-LMD$\uparrow$}
& \raisebox{-0.6ex}{I-PSNR/I-SSIM$\downarrow$} & \raisebox{-0.6ex}{FID$\uparrow$} & \raisebox{-0.6ex}{Sync$\downarrow$} & \raisebox{-0.6ex}{M-LMD$\uparrow$}
& \raisebox{-0.6ex}{I-PSNR/I-SSIM$\downarrow$} & \raisebox{-0.6ex}{FID$\uparrow$} & \raisebox{-0.6ex}{Sync$\downarrow$} & \raisebox{-0.6ex}{M-LMD$\uparrow$} \\
\midrule
AdvDm~\cite{advdm}
& 28.39/0.66 & 150.69 & 5.43 & 3.59
& 10.86/0.26 & 218.20 & 5.92 & 2.88
& 28.49/0.76 & 38.56 & 6.01 & 2.74
& 18.88/0.49 & 65.14 & 5.92 & \underline{3.43} \\
PhotoGuard~\cite{photoguard}
& 28.42/0.66 & 50.90 & 6.09 & 3.29
& 10.97/0.31 & 212.96 & 5.84 & 2.68
& \textbf{27.54}/\textbf{0.74} & \underline{40.66} & 5.94 & 2.85
& 18.89/\textbf{0.46} & 68.41 & 5.85 & 3.36 \\
Mist~\cite{mist}
& \textbf{27.70}/\textbf{0.64} & 147.98 & 5.64 & 3.93
& 10.82/0.28 & 216.38 & 5.71 & 3.00
& \underline{27.55}/\underline{0.75} & 38.55 & 6.12 & 2.89
& \underline{18.65}/\underline{0.47} & 70.32 & 5.86 & 3.36 \\
SDS(+)~\cite{sds}
& \underline{28.31}/0.67 & 134.52 & 5.69 & 3.67
& \underline{10.77}/\underline{0.25} & 218.10 & 5.86 & 2.88
& 28.47/0.76 & 37.60 & 6.05 & 2.86
& 18.69/0.48 & 65.31 & 5.95 & 3.31 \\
SDS(-)~\cite{sds}
& 28.97/\underline{0.65} & 44.25 & 5.94 & 3.38
& 10.88/0.31 & 200.62 & 5.90 & 2.77
& 28.23/\underline{0.75} & 38.55 & \underline{5.91} & 2.89
& 18.97/\underline{0.47} & 65.10 & 5.87 & 3.34 \\
Silencer-I~\cite{silence}
& 30.76/0.75 & 94.02 & 4.76 & 4.04
& 10.93/0.31 & 213.72 & 5.80 & 2.89
& 28.50/0.76 & 37.62 & 5.96 & 2.73
& 18.71/0.48 & 62.40 & \underline{5.83} & 3.33 \\
\rowcolor{lightblue}
\textbf{MIS}
& 30.29/0.73 & \underline{168.79} & \underline{3.24} & \underline{5.38}
& \textbf{8.84}/\textbf{0.19} & \underline{264.61} & \underline{5.61} & \underline{7.58}
& 28.32/\underline{0.75} & 37.63 & 5.92 & \underline{2.91}
& \textbf{18.47}/\underline{0.47} & \underline{71.73} & 5.89 & 3.42 \\
\rowcolor{lightblue}
\textbf{Ours}
& 30.29/0.73 & \textbf{170.54} & \textbf{0.90} & \textbf{6.16}
& \textbf{8.84}/\textbf{0.19} & \textbf{267.94} & \textbf{1.52} & \textbf{7.99}
& 28.32/\underline{0.75} & \textbf{42.44} & \textbf{1.92} & \textbf{4.90}
& \textbf{18.47}/\underline{0.47} & \textbf{76.45} & \textbf{1.60} & \textbf{4.97} \\
\bottomrule
\end{tabular}%
}
\label{tab:image-pure}
\end{table*}
\begin{table*}[t]
\centering
\mytablefont
\setlength{\tabcolsep}{2.0pt}
\renewcommand{\arraystretch}{1.0}
\caption{Quantitative comparison of different audio attack methods under four purifiers.}
\resizebox{\textwidth}{!}{%
\begin{tabular}{l|cccc|cccc|cccc|cccc}
\toprule
\multirow{2}{*}{Method}
& \multicolumn{4}{c|}{Spectral Gating~\cite{sgate}}
& \multicolumn{4}{c|}{Spectral Subtraction~\cite{ssub}}
& \multicolumn{4}{c|}{DiffWave~\cite{diffwave}}
& \multicolumn{4}{c}{WavePurifier~\cite{wavepure}} \\
\cline{2-17}
& \raisebox{-0.6ex}{SNR/PESQ$\downarrow$} & \raisebox{-0.6ex}{FID$\uparrow$} & \raisebox{-0.6ex}{Sync$\downarrow$} & \raisebox{-0.6ex}{M-LMD$\uparrow$}
& \raisebox{-0.6ex}{SNR/PESQ$\downarrow$} & \raisebox{-0.6ex}{FID$\uparrow$} & \raisebox{-0.6ex}{Sync$\downarrow$} & \raisebox{-0.6ex}{M-LMD$\uparrow$}
& \raisebox{-0.6ex}{SNR/PESQ$\downarrow$} & \raisebox{-0.6ex}{FID$\uparrow$} & \raisebox{-0.6ex}{Sync$\downarrow$} & \raisebox{-0.6ex}{M-LMD$\uparrow$}
& \raisebox{-0.6ex}{SNR/PESQ$\downarrow$} & \raisebox{-0.6ex}{FID$\uparrow$} & \raisebox{-0.6ex}{Sync$\downarrow$} & \raisebox{-0.6ex}{M-LMD$\uparrow$} \\
\midrule
FW-C\&W~\cite{whisper}
& \underline{2.45}/\underline{1.06} & 5.04 & 4.12 & 3.74
& \textbf{-2.94}/\underline{1.04} & 5.6 & 3.63 & 4.05
& \underline{5.95}/\underline{1.08} & 5.38 & 3.93 & 3.99
& \underline{1.07}/\underline{1.04} & 6.24 & 3.55 & 4.25 \\
FW-PGD~\cite{whisper}
& 2.97/1.12 & 4.18 & 5.07 & 3.03
& -2.79/1.17 & 4.53 & 4.92 & 3.43
& 12.05/1.39 & 4.5 & 4.74 & 3.27
& 1.20/1.11 & 4.35 & 4.96 & 3.29 \\
MW~\cite{mute}
& --- & \underline{6.13} & 5.03 & 3.30
& --- & \underline{6.85} & 5.91 & 3.30
& --- & \underline{6.86} & 4.80 & 3.73
& --- & \underline{6.82} & 5.35 & 3.52 \\
AA-C\&W~\cite{ASRAttacks}
& 2.71/1.13 & 4.43 & 5.05 & 3.04
& \underline{-2.85}/1.27 & 4.2 & 5.41 & 3.04
& 11.57/1.37 & 4.69 & 4.68 & 3.31
& 1.09/1.12 & 4.55 & 5.03 & 3.17 \\
AA-PGD~\cite{ASRAttacks} 
& \textbf{2.25}/\textbf{1.05} & 4.88 & \underline{3.33} & \underline{4.09}
& -2.81/\textbf{1.03} & 5.79 & \underline{2.66} & \underline{4.40}
& \textbf{2.42}/\textbf{1.02} & 5.88 & \underline{3.44} & \underline{4.18}
& \textbf{0.95}/\textbf{1.03} & 5.79 & \underline{2.75} & \underline{4.72} \\
\rowcolor{lightblue}
\textbf{CAF}
& 2.94/1.18 & 4.21 & 5.01 & 2.87
& -2.58/1.33 & 3.97 & 5.44 & 2.56
& 12.12/1.41 & 4.47 & 4.64 & 4.00
& 1.12/1.16 & 4.36 & 5.11 & 3.06 \\
\rowcolor{lightblue}
\textbf{Ours}
& 2.94/1.18 & \textbf{205.95} & \textbf{2.37} & \textbf{5.88}
& -2.58/1.33 & \textbf{205.31} & \textbf{2.56} & \textbf{5.73}
& 12.12/1.41 & \textbf{204.54} & \textbf{2.13} & \textbf{5.98}
& 1.12/1.16 & \textbf{204.80} & \textbf{2.40} & \textbf{5.23} \\
\bottomrule
\end{tabular}%
}
\label{tab:audio-pure}
\end{table*}

\begin{table}
\centering
\mytablefont
\setlength{\tabcolsep}{5.0pt}
\renewcommand{\arraystretch}{0.95}
\caption{Quantitative comparison of single-interval variants and MIS for the image-stream timestep setting.}
\begin{tabular}{l|ccccc}
\toprule
Interval & V-PSNR$\downarrow$ & V-SSIM$\downarrow$ & FID$\uparrow$ & Sync$\downarrow$ & M-LMD$\uparrow$ \\
\midrule
{[0,100]}    & \textbf{19.87} & \textbf{0.44} & \textbf{210.32} & \underline{3.19} & 5.42 \\
{[200,300]}  & 21.86 & 0.50 & 176.32 & 3.30 & \underline{5.46} \\
{[500,600]}  & 20.63 & 0.51 & 142.54 & 3.95 & 4.88 \\
{[700,800]}  & 20.68 & 0.51 & 154.30 & 4.71 & 4.21 \\
{[900,1000]} & 20.12 & 0.48 & 179.05 & 3.36 & 5.43 \\
\rowcolor{lightblue}
\textbf{MIS} & \underline{20.05} & \underline{0.46} & \underline{203.96} & \textbf{2.82} & \textbf{5.65} \\
\bottomrule
\end{tabular}
\label{tab:intervals_mis}
\end{table}

\begin{table}
\centering
\mytablefont
\setlength{\tabcolsep}{5.0pt}
\renewcommand{\arraystretch}{0.95}
\caption{Quantitative comparison of different layers within the same branch under the $[700,1000]$ interval. }
\begin{tabular}{l|ccccc}
\toprule
Layer & V-PSNR$\downarrow$ & V-SSIM$\downarrow$ & FID$\uparrow$ & Sync$\downarrow$ & M-LMD$\uparrow$ \\
\midrule
down\_0 & 23.10 & 0.74 & 6.94 & \underline{2.06} & 4.46 \\
mid\_0  & 22.84 & \underline{0.73} & 6.46 & 2.72 & \underline{4.47} \\
up\_1   & \textbf{22.49} & \textbf{0.72} & \underline{7.07} & 2.84 & 4.34 \\
\rowcolor{lightblue}
\textbf{CAF} & \underline{22.76} & \textbf{0.72} & \textbf{8.60} & \textbf{1.85} & \textbf{4.60} \\
\bottomrule
\end{tabular}
\label{tab:layer}
\end{table}

\section{Experiments}
  \subsection{Experimental Setup}
\subsubsection{Implementation Details}
We use the public Hallo~\cite{hallo} implementation at 25 FPS, with 16 kHz audio and $512\times512$ reference portraits. Hallo serves as the white-box victim model. All attacks run for 100 iterations. Image perturbations use an $\ell_\infty$ budget of $16/255$, consistent with all image baselines. Audio perturbations generated by CAF satisfy the peak-amplitude constraint
\begin{align}
dB_x(\delta)
=20\log_{10}\left(\frac{\|\delta\|_\infty}{\|x\|_\infty}\right)
\leq -30\,\mathrm{dB},
\end{align}
as defined in~\cite{asr4}. This optimization constraint differs from the energy-based SNR reported as an input-fidelity metric. The audio baselines retain their default parameter settings, with their iteration counts uniformly set to 100.

\textbf{Baselines and Datasets.}
Image baselines include AdvDM~\cite{advdm}, PhotoGuard~\cite{photoguard}, Mist~\cite{mist}, the SDS(+) and SDS(-) variants~\cite{sds}, and Silencer-I~\cite{silence}. Audio baselines include the C\&W and PGD variants of Fooling Whisper~\cite{whisper}, Muting Whisper~\cite{mute}, and the C\&W and PGD variants of ASRAdversarialAttacks~\cite{ASRAttacks}. We denote them FW-C\&W, FW-PGD, MW, AA-C\&W, and AA-PGD.

We construct two test protocols from three public datasets. The first pairs 50 CelebA-HQ~\cite{celea} portraits with 50 LibriSpeech~\cite{libri} audio clips. The second uses 50 HDTF~\cite{Flow} clips, taking the first frame of each clip as the reference portrait and retaining its original audio. Following Silencer~\cite{silence}, this evaluation scale is comparable to that commonly used in talking-head generation studies.

\begin{table}
\centering
\mytablefont
\setlength{\tabcolsep}{5.0pt}
\renewcommand{\arraystretch}{0.95}
\caption{Quantitative comparison across different timestep intervals for the {mid\_0\_lip} layer-branch unit.}
\begin{tabular}{l|ccccc}
\toprule
Interval & V-PSNR$\downarrow$ & V-SSIM$\downarrow$ & FID$\uparrow$ & Sync$\downarrow$ & M-LMD$\uparrow$ \\
\midrule
{[0,100]}    & 23.46 & 0.74 & 5.85 & 3.03 & 4.05 \\
{[400,600]}   & 23.45 & 0.74 & 6.24 & \underline{2.52} & 4.21 \\
{[900,1000] }& \textbf{22.71} & \underline{0.73} & \underline{6.52} & 2.53 & \underline{4.56} \\
\rowcolor{lightblue}
\textbf{CAF} & \underline{22.76} & \textbf{0.72} & \textbf{8.60} & \textbf{1.85} & \textbf{4.60} \\
\bottomrule
\end{tabular}
\label{tab:interval}
\end{table}
\subsubsection{Metrics}
Input fidelity is measured using I-PSNR and I-SSIM~\cite{ssim} for portraits and SNR and PESQ~\cite{pesq} for time-aligned audio. These audio metrics are omitted for prefix-based MW because prepending a segment breaks waveform alignment and makes pointwise comparison invalid. Output disruption is measured by V-PSNR, V-SSIM, and FID~\cite{fid} against videos generated from clean inputs. The first two quantify frame-level departure, while FID captures the distribution gap between clean and protected-input generations. SyncNet confidence~\cite{latentsync,sync1} measures lip--speech alignment, whereas M-LMD~\cite{ATVGnet} measures mouth-motion inconsistency. Stronger attacks lower V-PSNR, V-SSIM, and Sync while raising FID and M-LMD.

  \subsection{Privacy Protection}
Table~\ref{tab:all} shows that MIS and CAF disrupt different aspects of generation. MIS perturbs the reference portrait, which provides appearance and identity cues, and therefore causes larger changes in V-PSNR, V-SSIM, and FID, reaching FID 203.96/203.74 on CelebA-HQ--LibriSpeech/HDTF. CAF perturbs the driving audio, whose primary role is to guide facial motion. It consequently changes visual quality less but directly weakens local motion guidance, achieving the lowest Sync among audio baselines at 1.85/2.50. Combined, MIS and CAF yield FID 210.43/204.28, Sync 0.85/1.07, and M-LMD 6.26/3.68, jointly degrading appearance and speech-driven motion.

Despite this output disruption, the protected inputs remain perceptually close to the originals. Tables~\ref{tab:image-ae} and~\ref{tab:audio-ae} report MIS I-PSNR/I-SSIM of 29.56/0.69 and 29.59/0.66, and CAF SNR/PESQ of 24.86/1.63 and 26.53/2.45, on CelebA-HQ--LibriSpeech and HDTF, respectively. Fig.~\ref{fig:all-qual} shows the corresponding generated videos.
\subsection{Anti-Purification Experiments}

To test preprocessing defenses, we purify protected portraits or audio before generation. We evaluate JPEG~\cite{jpeg}, Resize~\cite{resize}, DiffPure~\cite{diffpure}, and DiffShortcut~\cite{DiffShortcut} for portraits, and Spectral Gating~\cite{sgate}, Spectral Subtraction~\cite{ssub}, DiffWave~\cite{diffwave}, and WavePurifier~\cite{wavepure} for audio. For single-stream evaluation, only the protected modality is purified. MIS pairs a purified protected portrait with clean audio, whereas CAF pairs a clean portrait with purified protected audio.

\textbf{Image purification.}
Robustness requires the purified portrait to remain distinct from the clean input, reflected by low I-PSNR/I-SSIM, while still yielding low Sync and high FID/M-LMD. As shown in Table~\ref{tab:image-pure}, MIS is strongest under JPEG and Resize. Under DiffPure and DiffShortcut, it is slightly weaker than some baselines, likely because those methods introduce larger facial distortions that purification does not fully remove.

\textbf{Audio purification.}
Table~\ref{tab:audio-pure} reports SNR/PESQ relative to clean audio and the output metrics after purification. Very noisy ASR attacks can appear robust when purification removes useful speech together with noise, passively lowering Sync or raising M-LMD by weakening mouth motion. The low SNR/PESQ values of FW-C\&W~\cite{whisper} and AA-PGD~\cite{ASRAttacks} in Table~\ref{tab:audio-ae} indicate this failure mode. CAF behaves more consistently relative to methods with more controlled distortion, such as FW-PGD~\cite{whisper} and AA-C\&W~\cite{ASRAttacks}.

\textbf{Mixed-stream evaluation.}
The Ours rows are mixed-stream evaluations rather than isolated tests that both streams survive purification. In Table~\ref{tab:image-pure}, a purified MIS portrait is paired with CAF audio, so gains over standalone CAF indicate residual MIS effects after image purification. In Table~\ref{tab:audio-pure}, a MIS portrait is paired with purified CAF audio, so lower Sync and higher M-LMD than standalone MIS indicate residual CAF effects after audio purification.

  \subsection{Ablation Study}
\textbf{MIS.}
MIS samples one timestep from each of $[0,100]$, $[100,200]$, $[300,400]$, and $[900,1000]$ per update, ensuring that every selected stage contributes to the update. Table~\ref{tab:intervals_mis} shows that no single interval dominates across metrics. High-noise timesteps affect global structure, intermediate intervals shape facial geometry, and low-noise timesteps refine lip details and textures. These complementary effects explain why MIS achieves stronger synchronization disruption, with Sync 2.82 and M-LMD 5.65, while retaining competitive visual degradation.

\textbf{CAF.}
Tables~\ref{tab:layer} and~\ref{tab:interval} isolate layer and timestep choices, respectively. The former fixes the interval and varies the U-Net layer, while the latter fixes a layer--branch unit and varies the interval. The resulting differences confirm that attention responses depend jointly on layer and denoising stage. Interval-specific target sets outperform the fixed selections and yield the best CAF result, with Sync 1.85 and FID 8.60.

\section{Limitation and Conclusion}
In this paper, we proposed SyncBreaker, a multimodal proactive protection framework for audio-driven talking-head generation. SyncBreaker combined image-stream MIS-based nullifying supervision with audio-stream CAF loss to jointly weaken speech-driven facial dynamics from both visual and acoustic conditioning pathways. Extensive experiments showed that the multimodal protective perturbations generated by our method effectively degraded facial dynamics, particularly audio-lip synchronization, while preserving the high perceptual quality of the protected inputs.

Our current study is limited to the white-box setting. Evaluating the transferability to unseen talking-head generation models in black-box scenarios remains an important direction for future work. We also plan to extend SyncBreaker to a wider range of portrait animation frameworks and more realistic deployment settings.
\bibliographystyle{plain}
\bibliography{sample-base}









\end{document}